\documentclass[runningheads]{llncs}

\usepackage{eccv}

\usepackage{eccvabbrv}

\usepackage{graphicx}
\usepackage{booktabs}

\usepackage[accsupp]{axessibility}  

\usepackage{hyperref}

\usepackage{orcidlink}

\begin{document}

\title{Cultural Heritage 3D Reconstruction with Diffusion Networks} 


\author{Pablo Jaramillo\inst{1}\orcidlink{0009-0006-6202-4907} \and
Ivan Sipiran\inst{1}\orcidlink{0000-0002-8766-3581}}

\authorrunning{Jaramillo and Sipiran}

\institute{Department of Computer Science, University of Chile, Chile
\email{pablo.jaramillo@ug.uchile.cl}\\ \email{isipiran@dcc.uchile.cl}
}

\maketitle

\begin{abstract}
  This article explores the use of recent generative AI algorithms for repairing cultural heritage objects, leveraging a conditional diffusion model designed to reconstruct 3D point clouds effectively. Our study evaluates the model's performance across general and cultural heritage-specific settings. Results indicate that, with considerations for object variability, the diffusion model can accurately reproduce cultural heritage geometries. Despite encountering challenges like data diversity and outlier sensitivity, the model demonstrates significant potential in artifact restoration research. This work lays groundwork for advancing restoration methodologies for ancient artifacts using AI technologies. \footnote{The dataset is available in: \url{https://github.com/PJaramilloV/Precolombian-Dataset}, and the code in \url{https://github.com/PJaramilloV/pcdiff-method}}
  \keywords{Automatic 3D Reconstruction \and Diffusion Models \and Deep Learning \and Point Clouds }
\end{abstract}

\section{Introduction}

The reconstruction of everyday items is a task as old as humanity itself and as diverse as any other technique. In recent decades, 3D printing has exploded in popularity, and with a vast and creative community, the technique and technology have advanced so quickly that on-site rapid construction of items ranging from whole products to replacements to tools is common in homes and industries. 

Despite the rapid growth in 3D generational technology, the design of these items, especially when requiring repairs of non-replaceable items, often involves a laborious manual task that requires previous experience with 3D modeling or CAD. Automated 3D reconstruction using Deep Learning models is a promising solution for this issue. The promise of this technique is simple: given a broken object, it can generate the shape required to restore it. However, the implementation is more elusive and subject to numerous research studies.

Recent progress in generative AI has opened the door to a myriad of applications in which it is increasingly feasible to generate information that fits well with the known data distribution. Consequently, the data generated are of high quality and very similar to the original data.  These characteristics make diffusion models an attractive alternative for the restoration of cultural heritage objects, mainly because high-quality and faithful restorations are required.  

This paper presents a method based on a diffusion model that conditions the generation of the missing parts of an object to the observed partial input. Our work was inspired by Friedrich et al.\cite{pcdmaig}, who proposed a diffusion method to generate cranial implants with high accuracy. We go a step further in our paper by first testing the generalizability of the diffusion model in~\cite{pcdmaig} and then adapting a specific methodology to train a model to repair archaeological parts.

The contributions of this study are twofold. First, we evaluate the capability of the diffusion model for the restoration of general objects. This evaluation seeks to determine the generalization capabilities of diffusion models on point clouds. Second, we adopt a conditional diffusion model that learns to generate plausible missing parts in cultural heritage objects. To achieve this goal, we devise a specific methodology that teaches the model to predict the missing geometry with high accuracy. 

The rest of the paper is organized as follows. Section~\ref{Sec:related} discusses related works. Section~\ref{Sec:Background}  introduces the background of the conditional diffusion model for point-cloud completion. Section~\ref{Sec:General} evaluates the diffusion model by using a general setup. Section~\ref{Sec:CH} focuses on the reconstruction of cultural heritage objects. Section~\ref{Sec:Limitations} presents the limitations of the proposal. Finally, Section~\ref{Sec:Conclusions} concludes the paper.

\section{Related Work}
\label{Sec:related}

Interest in three-dimensional reconstruction has increased significantly over the past several decades, resulting in the creation of numerous works that employ a diverse range of techniques and architectures. Each of these approaches offers unique advantages and characteristics that can be utilized to achieve specific objectives.

Examples of the architectures used are Generative Adversarial Networks \cite{ORGANHermozaSipiran, iberianVoxel, AVRGAN2020}, Occupancy Networks \cite{lamb2022deepjoin,lamb2022deepmend,lamb2022mendnet}, autoencoders \cite{cranialEncoder, park2019deepsdf,rdap}, and, more recently, diffusion models\cite{pointvoxeldiffusion, pcdmaig}. 

By examining the results and underlying techniques of these studies, we can identify opportunities for further exploration. For instance, models that employ voxel representations~\cite{ORGANHermozaSipiran,iberianVoxel} have demonstrated effectiveness; however, achieving a practical resolution for general 3D reconstruction remains a challenge. Other promising models alleviate the need for computationally expensive voxel representations using point clouds as a compact representation~\cite{park2019deepsdf,pointvoxeldiffusion,rdap}. Nevertheless, achieving high-quality results remains a challenge because of the arbitrary sampling of points in 3D space.  

More recently, reconstruction has been formulated as a completion task, where a model learns to predict a complete object given a partial observation. Diffusion models have also been used to address this problem. For example, some methods explore alternative representations, such as signed distance functions, along with diffusion modeling~\cite{sdfusion2023,diffcomplete2023}. There are also methods that explore a combination of diffusion models with text-driven completion~\cite{sds2023}. Some approaches learn to generate coarse point clouds that are subsequently refined~\cite{pdr2022}. Furthermore, the completion task in point clouds has been formulated as an algebraic problem that can be solved in combination with diffusion~\cite{nullspace2023} or as a rotation equivariance problem that requires specific architectures~\cite{rotequi2024}. Diffusion models have also been successfully applied for the generation of cranial implants~\cite{pcdmaig}.

\section{Background}
\label{Sec:Background}

In this section, we describe the details on conditional completion with diffusion models~\cite{pcdmaig}. 

\subsection{Completion with Diffusion Models}

The task of reconstruction in the point cloud space with a diffusion model involves a conditional generation process with a forward and backward pass through a chain of probabilistic changes, inspired by  \cite{pointvoxeldiffusion}, where the cloud of $N$ points to be repaired $\boldsymbol{c}_0 \in \mathbb{R}^{N\times 3}$ is the conditional basis for generating a cloud $x_0$ that contains all points from $c_0$ plus $M$ corresponding to the repair. During training, ground-truth repair is necessary to obtain cloud $\tilde{x}_0 \in \mathbb{R}^{M\times 3}$ thus, $x_0 = (\tilde{x}_0, \boldsymbol{c}_0)$. Both $N$ and $M$ are hyperparameters of the model chosen by the desired level of quality, and relative volume of the complete shape that a repair is expected to take. With this, we can define the conditional forward diffusion process modeled as a Markov chain, where there is a sequence of $T$ steps. Then, for each step, small additions of noise were made to $\tilde{x}_0$, gradually shifting the positions of its points while keeping $\boldsymbol{c}_0$ unaffected. This results in clouds $x_i \in \left \{ x_i,\  x_i = (\tilde{x_i}, \boldsymbol{c}_0),\ \forall \ i \in \mathbb{N}_0,  i \leq T \right \}$ and fits a distribution:

\begin{equation}
    q(\tilde{x}_{0:T}) = q(\tilde{x}_0) \prod_{t=1}^{T}q(\tilde{x}_t|\tilde{x}_{t-1})
\end{equation} 

The transitions consist of a parameterized Gaussian distribution that can have a set variance schedule $\beta_1, ..., \beta_T$ to control the rate of diffusion.

\begin{equation}
    q(\tilde{x}_t|\tilde{x}_{t-1}):= \mathcal{N}\left ( \sqrt{1-\beta_t} \tilde{x}_{t-1}, \beta_t \boldsymbol{I} \right)
\end{equation} 

Then the backward or ``reverse'' diffusion process undoes the formerly described process by removing noise from $\tilde{x}\sim \mathcal{N}(0,\boldsymbol{I})$ in a similar fashion under another Gaussian distribution as follows:

\begin{equation}
    p_{\theta}(\tilde{x}_{0:T}) = p(\tilde{x}_0) \prod_{t=1}^{T}p_{\theta}(\tilde{x}_{t-1}|\tilde{x}_t ,\boldsymbol{c}_0)
\end{equation}
\begin{equation}
    p_{\theta}(\tilde{x}_{t-1}|\tilde{x}_t ,\boldsymbol{c}_0):= \mathcal{N}\left ( \mu_{\theta}(\tilde{x}_{t},\boldsymbol{c}_0 , t), \beta_t \boldsymbol{I} \right)
\end{equation}

The goal of this model is to make the network $\epsilon_{\theta}$ predict the noise to be removed from a noisy input $\tilde{x}_t$, which is formalized as $\epsilon_{\theta}(\tilde{x}_t,\boldsymbol{c}_0,t)$ for a time step $t < T$. The training loss of the model is a mean-squared error optimization calculated for a random time step, as follows:

\begin{equation}
    \mathcal{L}_t  = \left\| \epsilon - \epsilon_{\theta}(\tilde{x}_t, \boldsymbol{c}_0 , t)\right\|^2, \  \textrm{where} \ \tilde{x}_t = \sqrt{\tilde{\alpha}_t}\tilde{x}_0 + \sqrt{1-\tilde{\alpha}}\epsilon
\end{equation}

where $\epsilon$ is a Gaussian of the variance $\boldsymbol{I}$, $\alpha_t = 1 -\beta_t$ and $\tilde{a}_t = \prod_{s=1}^{t}\alpha_s$. Once the network is trained, we can use a Gaussian distribution of noise to make $x_T = (\tilde{x}_T, \boldsymbol{c}_0)$, which can be passed through the backward process to center the noise into a usable repair.

\begin{equation}
    \tilde{x}_{t-1} = \frac{1}{\sqrt{\alpha_t}}\left ( \tilde{x}_t - \frac{1-\alpha_t}{\sqrt{1 - \tilde{\alpha}_t}} \epsilon_{\theta}(\tilde{x}_t, \boldsymbol{c}_0, t)  \right ) + \sqrt{\beta_t}z
\end{equation}

where $z \sim \mathcal{N}(0,\boldsymbol{I})$ until $t=1$. The network of \cite{pcdmaig} corresponds to a traditional conditional diffusion model was inspired by the propositions from \cite{pointvoxeldiffusion} and utilizes an architecture based on \cite{pointvoxel,pointnetpp}. The model consists of a CNN with 4 PointNet Set Abstraction layers through which the point clouds go through followed by 4 Feature Propagation layers, and later a multi-layer perceptron. The output of those blocks are then fed to the explained Gaussian Diffusion model. The PointNet blocks utilize PointVoxel convolutional modules.

\subsubsection{Data}
It is worth noting that the original work~\cite{pcdmaig} reports the usage of the publicly available parts of the SkullBreak and SkullFix datasets from \cite{Skulls}, which is a selection of the CQ500 dataset from \cite{CQ500}. The data consist of 114 and 100 CT scans of human skulls artificially broken in five different classes of trauma or surgery and sorted such that for each class of injury, every skull has a complete and healthy representation, a broken representation, and a ground truth implant, resulting in 570 triplets for SkullBreak and 500 for SkullFix, split in a cumulative 75-25 split for training the network responsible for the task of completion, which is such that all versions of each skull are set destined for one part of the split.

\section{Generalization Experiment}
\label{Sec:General}

The methodology and design behind the model are sound and do not provide indications as to why it would not perform in other scenarios. We then decided to try out the model using other datasets with more generic objects. 

As a starting point, we decided to use the geometric break dataset from \cite{lamb2022deepjoin} which contains a variety of item classes similar to other popular datasets. These objects feature breakages with a degree of realism that, while not perfect, deliver fair representations of the surface patterns that can be seen in real-life breakages. 

In particular, we tried out the bottle, cup, pot, and vase classes because they can represent practical, easy to conceptualize applications where 3D reconstruction can be of use. These have 447, 178, 500, and 500 different objects in obj mesh files, each broken in different ways such that they totaled 1274, 1559, 1377, and 1377 triplets of complete-broken-repair, respectively. This dataset was ideal for testing the implementation, given that PCDiff's model~\cite{pcdmaig} requires these three files for each object.

The files were serialized and organized into three different directories per data class: complete, broken, and repair, each of which had the corresponding file. The project was altered to fit this simplified and easy-to-match structure, with no notion of classes.

The model was independently trained for each class derived from Geometric Breaks to accommodate the substantial diversity within the dataset. While human skulls exhibit some differences, they generally share similar features, with variations being relatively minor compared to the significant differences among items from the aforementioned classes.

The training process utilized a 70-30 data split and a downscaling of the original hyperparameters. The abstraction layer sizes were set to 2048, 1024, 512, and 160, while the epoch count was reduced to a third of the original, and the point count for both inputs and implants was initially set to a tenth of that specified in the original paper. This resulted in 3080 points for the input broken object, 308 points for reconstructing the missing part, and 5000 epochs of training, aiming to provide proof of concept results.

During the experimental phase, the hyperparameters were iteratively adjusted based on visual analysis of the results until they reached a stable state. The final values were set to 8192 points for the broken object, 820 points for the repair, and 500 epochs of training. The abstraction layer sizes were maintained at the initially stated dimensions.

\subsection{Evaluating Results}

To evaluate the results, we devised a metric to determine the proximity of a given repair to an optimal repair. This metric involves calculating the Chamfer and Haussdorf distances \cite{point-cloud-utils} between our results and the ground truth, normalized by the respective distance between the ground truth itself. Specifically, for a given repair shape \( S \), our calculated shape \( S_r \), and its ground truth \( S_g \), where the ground truth consists of \( N \) points and the calculated shape consists of \( M \) points with \( N > M \), we use a method for random down sampling \( R \) that reduces a shape to a target number of points. For a distance metric \( D \), we calculate the distance factor \( F \) as follows:

$$F_D = \frac{D(S_r, R(S_g, M))}{D(R(S_g, M),R(S_g, M))}$$

The concept involves using a pair of random downsamplings of the ground truth to approximate the realistic lower bound for the distance between a generated repair shape and its ground truth counterpart. Consequently, the factor derived from the division in the aforementioned formula serves as an indicator of how close the generated shape is to an ideal reconstruction, assuming a downsample of the given ground truth is sufficiently useful. Naturally, this metric is designed such that a lower value indicates higher precision and quality of the repair. It is important to note that this metric is dependent on the data and the number of points present in the point cloud. Therefore, comparisons are only valid when all point clouds involved either have the same number of points or are downsampled to the same number of points.

This approach has an advantage over more common methods which perform up-scaling before evaluating because the quality of the repair remains dependent solely on the reconstruction method, and as such is sensitive to changes in small details. Consequently, if the downsampling method can faithfully represent finer details of the original object then this behavior rewards high degrees of precision fitting of finer details as those desired in archaeological reconstruction.

The previously described calculation was performed for all instances present in the test partition of the created dataset, and its statistical metrics were calculated later. We refer to the ``Chamfer Distance Factor'' as CDF and ``Hausdorff Distance Factor'' as HDF, the application of the proposed metric for those respective distances.

Table~\ref{Tab:general} shows the statistics of the described factors on the tested classes of Geometric Breaks with 820 points on the repair pieces.

\begin{table}[]
\centering
\begin{tabular}{l|ccccc|ccccc}
\textbf{PCDiff on} &      &        & \textbf{CDF} &      &       &      &       & \textbf{HDF} &      &       \\
\textbf{Dataset}   & min  & max    & avg          & med  & std   & min  & max   & avg          & med  & std   \\ \hline

GB-Bottles         & 1.08 & 95.89  & 5.15         & 1.77 & 12.56 & 0.35 & 70.52 & 6.08         & 2.37 & 9.40  \\
GB-Cups            & 1.01 & 99.86  & 3.13         & 1.54 & 8.35  & 0.71 & 48.91 & 3.73         & 4.45 & 2.29  \\
GB-Pots            & 1.11 & 230.97 & 18.61        & 3.49 & 32.66 & 0.82 & 94.08 & 13.76        & 8.12 & 15.13 \\
GB-Vases           & 1.26 & 121.57 & 15.58        & 2.56 & 25.56 & 1.36 & 57.81 & 12.79        & 6.93 & 6.93 
\end{tabular}
\caption{Results of applying ~\cite{pcdmaig} in a general setting. We report the ninimum, maximum, average, median and standard deviation for every experiment.}
\label{Tab:general}
\end{table}

\begin{figure}[ht!]
  \centering
  \includegraphics[width=.7\linewidth]{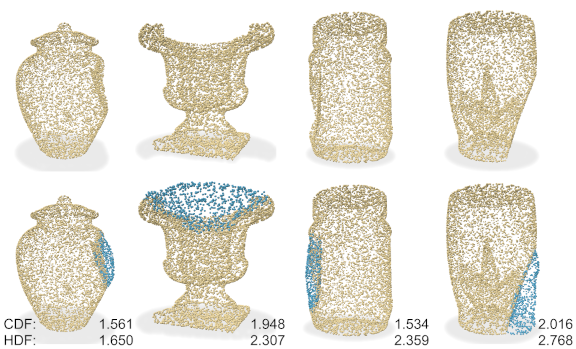}
  \caption{Examples of repaired pots. Top row displays the input, while the bottom row shows the composite of the input with the generated repair. All repairs have their according distance factors.}
  \label{fig:pot}
\end{figure}

\begin{figure}[ht!]
  \centering
  \includegraphics[width=0.7\linewidth]{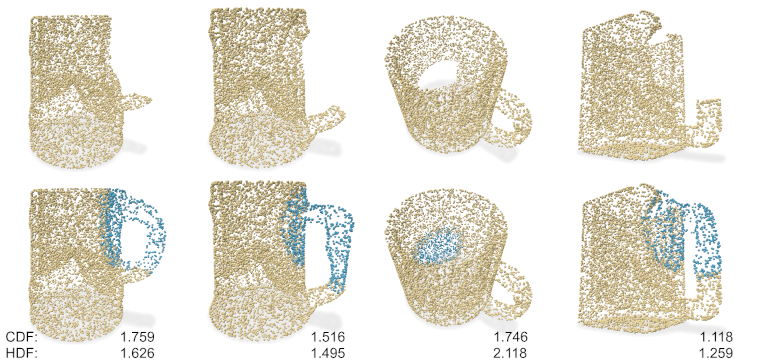}
  \caption{Examples of repaired cups. Top row displays the input, while the bottom row shows the composite of the input with the generated repair. All repairs have their according distance factors.}
  \label{fig:cup}
\end{figure}

\begin{figure}[ht!]
  \centering
  \includegraphics[width=.7\linewidth]{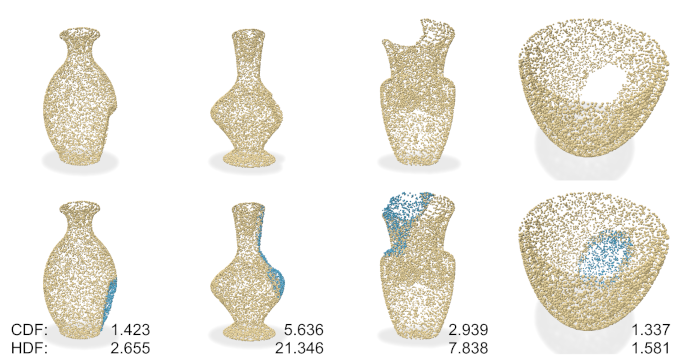}
  \caption{Examples of repaired vases. Top row displays the input, while the bottom row shows the composite of the input with the generated repair. All repairs have their according distance factors.}
  \label{fig:vase}
\end{figure}

\begin{figure}[ht!]
  \centering
  \includegraphics[width=0.5\linewidth]{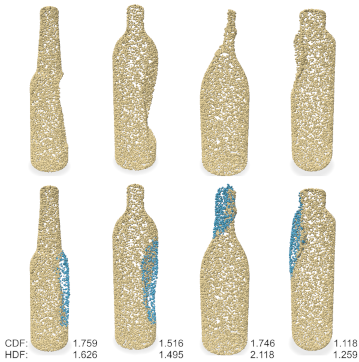}
  \caption{Examples of repaired bottles. Top row displays the input, while the bottom row shows the composite of the input with the generated repair. All repairs have their according distance factors.}
  \label{fig:bottle}
\end{figure}

Some of the better repairs generated by the network can be seen in Figures~\ref{fig:pot},\ref{fig:cup},\ref{fig:vase} and \ref{fig:bottle}, with the input showed independently on the top rows and the combined input and repair on the bottom row.



The first experiment shows the ability of the diffusion model to complete objects of various classes. Its better performance in classes such as bottles and cups can be attributed to the low variability of the objects in those classes, in contrast to the pot and vase classes, where the completion task is more challenging. Nevertheless, these results indicate that this conditional model is capable of good reconstruction when trained effectively.

\section{Reconstructing Archaeological Artifacts}
\label{Sec:CH}

Given the insights obtained from the experiments with Geometric Breaks, we decided to perform another experiment to see how the model would compare against related work. The field chosen for this comparison was archaeological reconstruction.

The particular application needs to be specific because of the amount of diversity in the data that the model can handle. To limit the scope of the application such that it could be practical, we drew inspiration from the Data-Driven Restoration of Digital Archaeological Pottery (DRDAP) \cite{rdap} to set the scope to restore the lower parts of containers from a specific cultural background, particularly those corresponding to the pre-Colombian South American context in the task of reconstructing missing bottoms due to scan failure.

\subsection{Dataset creation}

To implement the comparison, it was necessary to create a dataset that would approximate the variability of the already proven data. To this end, we compiled data from the Native American class from the Cultural Heritage dataset~\cite{ch_dataset}, the bowl class from ModelNet40~\cite{modelnet2015}, and added some bowls from ShapeNet~\cite{Shapenet2015} that were not already present in ModelNet-40. 

The process of performing this final addition consisted of rendering and reviewing all the objects from the datasets for manual acceptance into the dataset based on the criteria of resemblance to artifacts present in the testing of \cite{rdap}. To add data from ShapeNet not present in ModelNet-40, we obtained point cloud representations by performing the same Poisson Disk Sampling from PCDiff on all objects, normalized them, and calculated the chamfer distance for each object pair between the datasets. The top five matches for each ShapeNet bowl were then reviewed to check if they were indeed duplicates. In addition, we included the scanned artifacts from \cite{rdap} in the dataset as strictly test-only shapes because they do not have ground-truth restoration. Finally, the dataset had 253 bowls, bowl-like, and vase shapes.

\subsection{Data Augmentation}

To create the broken and repair representations, we performed straight cuts on the point clouds at a random height from the bottom 18 \% to 22\% of the object's height after a random rotation of less than 30° on the X and Z axes, where the coordinates have Y in the upward direction. The breakage was repeated four times per object to augment the data, resulting in 976 complete broken repair triplets. 

The model was trained on a 70-30 split of the created augmented dataset, similar to the original work, with all augmented data points for every object belonging to the same part of the split, meaning that the model testing was performed without seeing any variant of a test object in its training.

\subsection{Training}

The experiment was performed with the values arrived at with the classes from Geometric Breaks leading; later, the experiment was repeated with double the number of points and increased epoch count to 4000 epochs to more closely approximate the level of refinement of the PCDiff training.

\subsection{Results and comparison against DRDAP}

To compare the results we chose to train the publicly available implementation of 
DRDAP\cite{rdap} using a compiled dataset. This choice was made because we found it valuable to compare with existing related work in the chosen application field, its public availability, and the fact that it works with point cloud data. We used the same technique to create the distance factor as we described for the tests on general shapes. The following is a comparison over our compiled dataset's repair shapes, with 1638 points each between DRDAP and our application of PCDiff, both trained from scratch.

\begin{table}[]
\centering
\begin{tabular}{l|ccccc|ccccc}
\textbf{Model} &     &      & \textbf{CDF} &       &     &     &      & \textbf{HDF} &      &     \\
               & min & max  & avg          & med   & std & min & max  & avg          & med  & std \\ \hline
DRDAP\cite{rdap}&1.27 &77.70 & 10.27        & 7.07  &9.74 &1.53 &37.57 & 10.71        &10.15 & 5.12 \\
\textbf{PCDiff}  & \textbf{0.99} & \textbf{7.31} & \textbf{1.49}         & \textbf{1.33}  &\textbf{0.57} &\textbf{0.33} &\textbf{17.02} &  \textbf{1.43}        & \textbf{1.33} & \textbf{1.43} 
\end{tabular}
\caption{Completion evaluation with respect to previous methods.}
\label{table:pcdiffvdrdap}
\end{table}

\begin{figure}
  \centering
  \includegraphics[width=.9\linewidth]{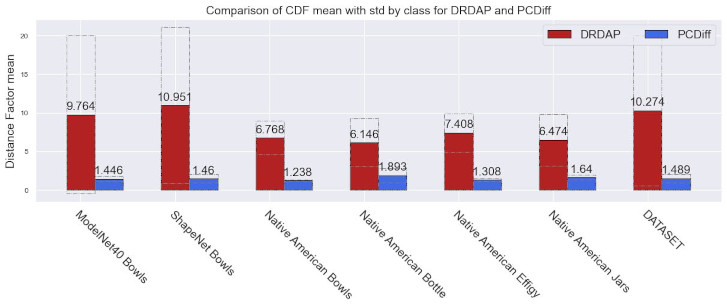}
  \caption{Bar plot of the CDF results presented in Table~\ref{table:pcdiffvdrdap} separated by data class of the compiled dataset. Dashed bars represent 1 standard deviation in both directions.}
  \label{fig:cdfplot}
\end{figure}%
\begin{figure}
  \centering
  \includegraphics[width=.9\linewidth]{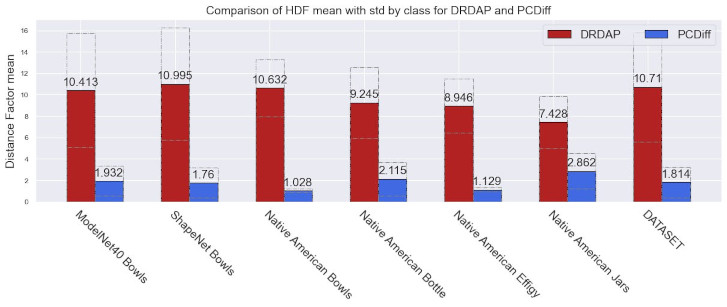}
  \caption{Bar plot of the HDF results presented in Table~\ref{table:pcdiffvdrdap} separated by data class of the compiled dataset. Dashed bars represent 1 standard deviation in both directions.}
  \label{fig:hdfplot}
\end{figure}%

\begin{figure}
  \centering
  \includegraphics[width=.7\linewidth]{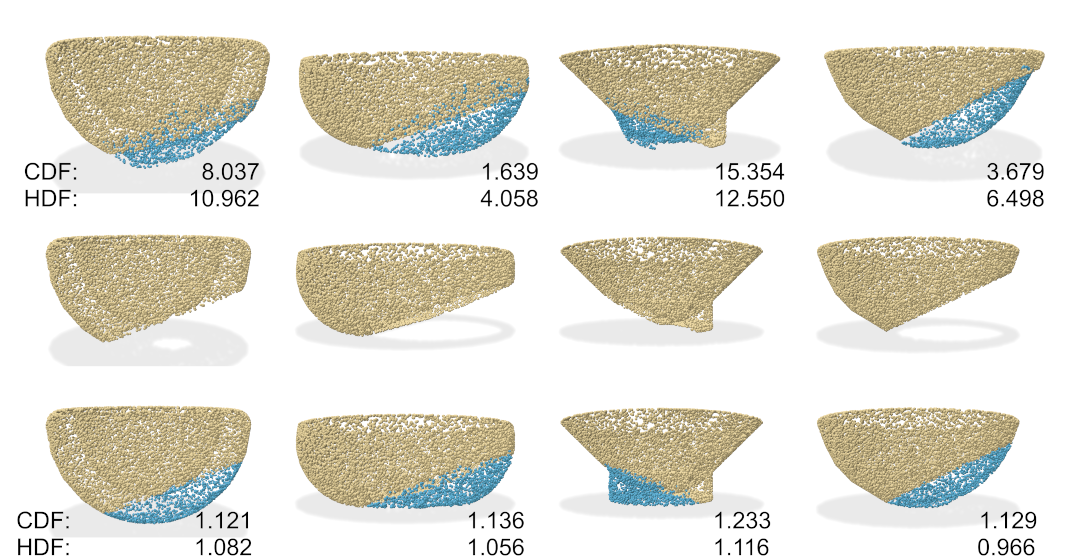}
  \caption{Comparison between the results from DRDAP and PCDiff over the ModelNet 40 clouds. Top row has the composite clouds of the repairs by DRDAP. Bottom row has the composite clouds of the repairs by PCDiff. Isolated input is shown on middle row. All repairs have their corresponding distance factors.}
  \label{fig:md_comp}
\end{figure}

\begin{figure}
  \centering
  \includegraphics[width=0.7\linewidth]{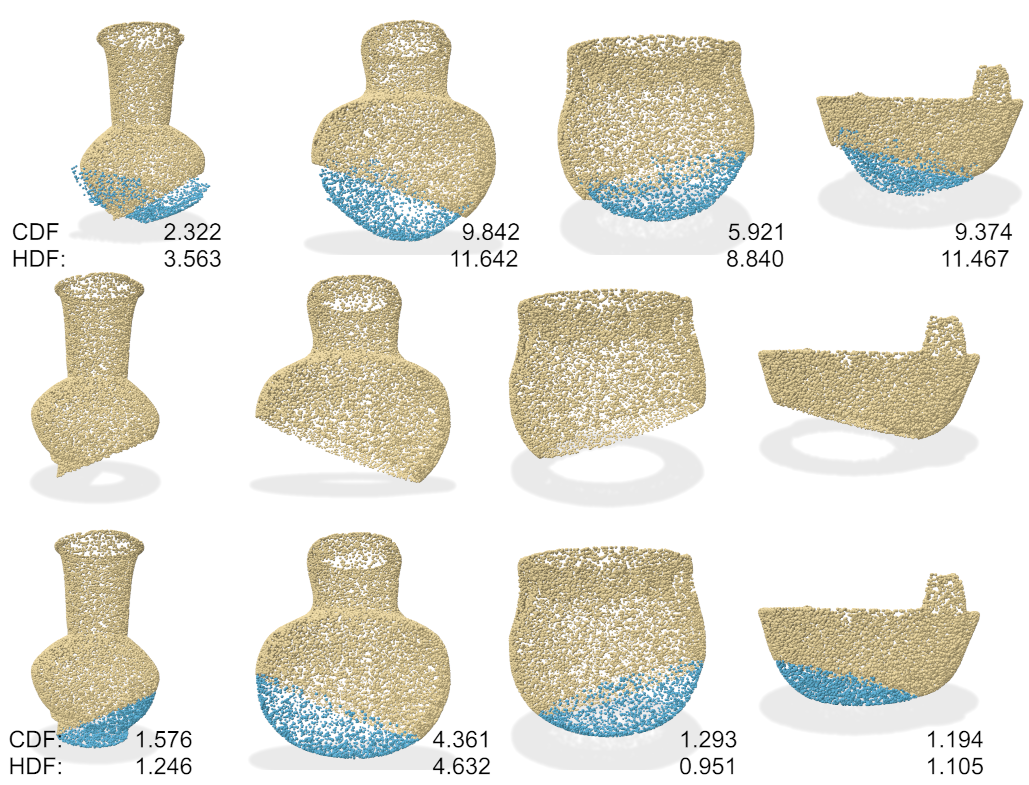}
  \caption{Comparison between the results from DRDAP and PCDiff over the Native American Pottery clouds. Top row has the composite clouds of the repairs by DRDAP. Bottom row has the composite clouds of the repairs by PCDiff. Isolated input is shown on middle row. All repairs have their corresponding distance factors.}
  \label{fig:na_comp}
\end{figure}

\begin{figure}[!h]
\centering
  \includegraphics[width=0.9\linewidth]{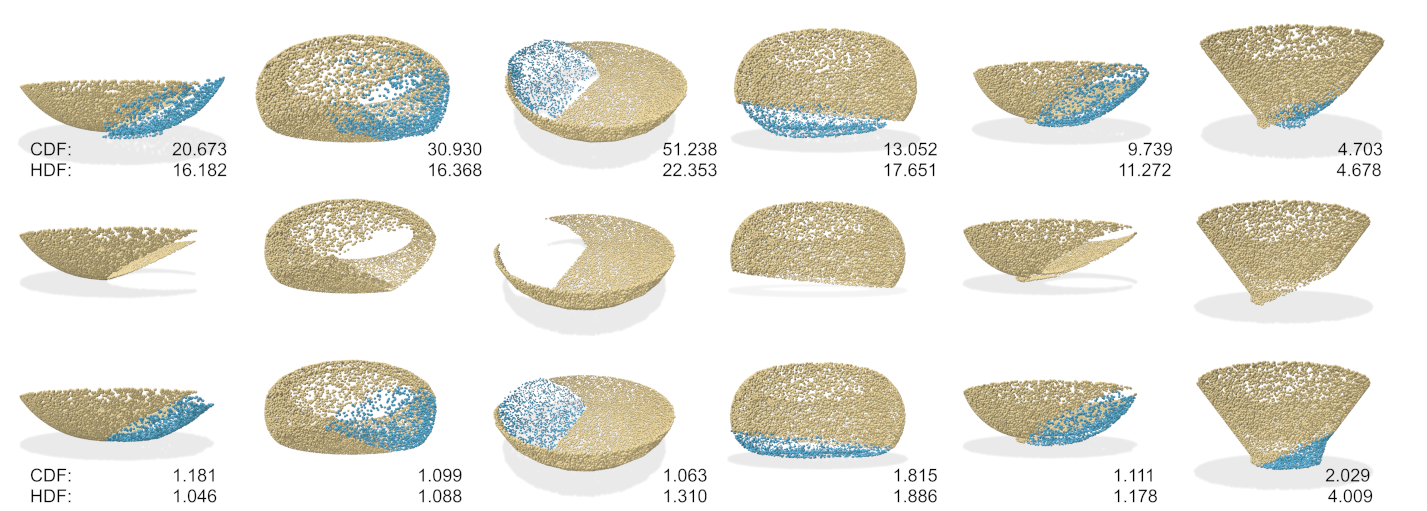}
\caption{Comparison between the results from DRDAP and PCDiff over the ShapeNet clouds. Top row has the composite clouds of the repairs by DRDAP. Bottom row has the composite clouds of the repairs by PCDiff. Isolated input is shown on middle row. All repairs have their corresponding distance factors.}
\label{fig:sn_comp}
\end{figure}

We can clearly see from the Table~\ref{table:pcdiffvdrdap} that PCDiff outperforms DRDAP substantially and achieves very low factors in both distance metrics, showing encouraging results. We include a group of comparison figures where some objects from the dataset are shown, see \ref{fig:md_comp}, \ref{fig:na_comp} and \ref{fig:sn_comp}.


\section{Limitations}
\label{Sec:Limitations}

When comparing the distance factors calculated for PCDiff, a substantial variation depending on the dataset is evident, highlighting potential limitations of the model.

The principal limitation of this model is its ability to handle diverse data. This limitation has not been formalized into a quantifiable expression and remains largely intuitive, based on the researcher's experience. This means as well that any fine tuning of the model should be an expansion or reinforcement of a desired quality in a given application. A secondary limitation arises when dealing with slim and narrow shapes that contrast sharply with broader main features; however, the model performs well if trained with data tailored to such features. These problems suggest that this model is not suited to reliably repair shapes with significant differences from the training data. 

Additional limitations become apparent upon reviewing the experimental results. The model is highly susceptible to outliers—data points that contribute no useful information. Characteristics of these detrimental objects include non-contiguous geometry or shapes irrelevant to the goal.

\begin{figure}
  \centering
  \includegraphics[width=.9\linewidth]{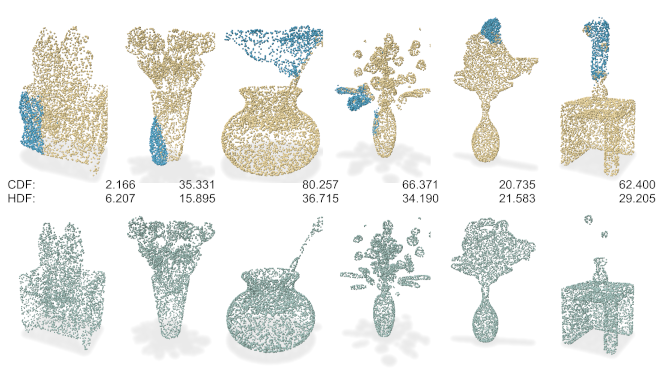}
\caption{Composite clouds of harmful data repaired by PCDiff compared against the original complete object. Distance factors included for reference.}
\label{fig:test}
\end{figure}

In summary, for the model to perform effectively, the dataset must be meticulously curated and purpose-driven. This was not the case for the Geometric Breaks data classes, where problematic shapes, as described, were identified during the review of results. Figure~\ref{fig:test} illustrates some flawed and erroneous repairs on objects that should not have been included in the dataset, as they contain geometry irrelevant to the data class. 

The oversight of including non useful objects negatively impacted the model. However, despite these issues, the model still managed to produce notable results.

In our opinion, data as the illustrated in said figure prevents convergence by adding unnecessary complexity to the general distribution of points representing the class. Contrasting with the archaeological reconstruction experiment, where the training data was a faithful representation of the test, the difference in performance shows how the quality is enhanced as the data distribution of the training approaches the test.

\section{Conclusions}
\label{Sec:Conclusions}

In this article, we address the challenge of repairing cultural heritage objects using recent generative AI algorithms that have demonstrated impressive results in other domains. We introduce the application of a conditional diffusion model capable of effectively reconstructing 3D point clouds. Our contribution is twofold. First, we experiment with a general setting to analyze the model's capabilities. Second, we evaluate the performance in the domain of cultural heritage. Our results indicate that, under certain assumptions regarding the variability of the objects, the diffusion model can successfully recreate the geometry of cultural heritage models. We believe our research can serve as a foundation for developing new and improved methods for the restoration of ancient artifacts.

\section*{Acknowledgements}
This work was supported by ANID Chile—Research Initiation Program—Grant N° 11220211 and National Center for Artificial Intelligence CENIA FB210017, Basal ANID, Chile.

\bibliographystyle{splncs04}
\bibliography{main}
\end{document}